# Ego-motion Estimation Based on Fusion of Images and Events

Liren Yang

Nanjing Institute of Intelligent Technology, Nanjing, China

*Abstract*—Event camera is a novel bio-inspired vision sensor that outputs event stream. In this paper, we propose a novel data fusion algorithm called EAS to fuse conventional intensity images with the event stream. The fusion result is applied to some ego-motion estimation frameworks, and is evaluated on a public dataset acquired in dim scenes. In our 3-DoF rotation estimation framework, EAS achieves the highest estimation accuracy among intensity images and representations of events including event slice, TS and SITS. Compared with original images, EAS reduces the average APE by 69%, benefiting from the inclusion of more features for tracking. The result shows that our algorithm effectively leverages the high dynamic range of event cameras to improve the performance of the ego-motion estimation framework based on optical flow tracking in difficult illumination conditions.

## 1. Introduction

Ego-motion estimation is that an object equipped with sensors estimate its own motion in a rigid scene, which holds great significance for autonomous robot navigation, self-driving cars and augmented reality. Nowadays, there are varieties of solutions for ego-motion estimation, involving different types of sensors like camera [1-3], radar [4-5], IMU [6-7], etc. For visual solutions, which are also referred as VO (visual odometry), the quality of acquired images has vital influence on the estimation accuracy. Therefore, how to ensure that it works well in nonideal illumination conditions is a major difficulty faced by VO.

Event camera is a novel bio-inspired vision sensor which only records changes of light intensity. Compared with traditional frame-based cameras, it has advantages of high dynamic range (130 dB versus 60 dB) and low latency. In addition, it won't suffer from motion blur on account of the microsecond temporal resolution. Because of these outstanding advantages, event camera has a huge potential to break through the bottleneck in current computer vision application. However, specific algorithms should be developed to apply it to related applications due to the unique form of its output. Commonly used event cameras include DVS, ATIS and DAVIS [8].

In this paper, in order to take advantage of the high dynamic range of event cameras, we propose a novel data fusion algorithm called EAS (Events Aggregation and Superimposition), which fuses conventional intensity images with the event stream output by event cameras. Then we take the fusion results as inputs to a visual 3-DoF rotation estimation framework and the monocular ORB-SLAM2 [9] respectively to evaluate the performance of EAS in ego-motion estimation through two dim sequences in a public dataset [10]. In the experiment, we compare it with original images, enhanced images, and some representations of events, including event slice, TS (time surface) [11], SITF (Speed Invariant Time Surface) [12].

In 3-DoF rotation estimation task, EAS has the lowest average APE (Absolute Pose Error), which is 69% lower than that of the original images. In 6-DoF pose estimation using ORB-SLAM2, because ORB features in EAS are not stable and tend to be mismatched, it doesn't work better. Experiment results show that proposed fusion algorithm can leverage the high dynamic range of event camera to obtain images with higher quality in difficult illumination conditions, and improve the performance of the visual ego-motion estimation frameworks based on optical flow tracking.

The remainder of the paper is organized as follows: Section 2 introduces related work about DVS involved in ego-motion estimation; Section 3 presents the data fusion method; Section 4 analyzes experiment results and evaluates the performance of the algorithm; Section 5 goes to conclusions of the paper.

## 2. Related Work

Since the output of event cameras is asynchronous event stream that completely differs from conventional intensity images, many specific event-based ego-motion estimation methods have been developed.

Some studies have focused on methods using event-based motion compensation. Guillermo G et al. [13] estimated the rotational motion of an event camera by maximizing contrast of accumulated events. Xu J et al. [14] directly aligned the curved event trajectories with time-varying motion parameters and introduced a smooth constraint in image energy. Nunes U M et al. [15] proposed an Entropy Minimization framework, extending the event-based motion compensation algorithms applied in motion estimation. For maximizing contrast, Wang Y et al. [16] maximized the contrast in a volumetric ray density field to estimate the motion of event camera mounted on AGV. Liu D et al. [17] estimated the rotational motion by spatiotemporal registration which has better performance than contrast maximization. Neural networks have also been applied in this field. Nguyen A et al. [18] introduced a Stacked Spatial LSTM network to learn the 6-DoF pose of event camera in real time. Zhu A Z et al. [19] and Ye C et al. [20] both proposed a framework based on unsupervised learning to estimate optical flow, depth and ego-motion with the input of events.

Some event-based SLAM frameworks have been proposed as well. Rebecq H et al. [21] and Zhou Y et al. [22] presented an event-based monocular and stereo visual odometry system respectively. Jiao J et al. [23] compared two image-type representations of events used in event-based SLAM and proposed a general strategy to switch proper representations online. Besides, some other sensors have been introduced and combined with event cameras for ego-motion estimation. Zuo Y F et al. [24] presented a novel real-time stereo visual odometry combining event cameras and RGB-D cameras, which improved the performance of SLAM in challenging lighting conditions. Le Gentil C et al. [25] designed and implemented an IMU-DVS odometry framework that combines IMU with event cameras.

## 3. Method

EAS is implemented in two steps: aggregate events into event slices; superimpose corresponding event slices on intensity images.

### A. Aggregate events

The output of event cameras is event stream, and the $i^{th}$ event can be represented by a 4-dimentional vector $e_i = (x_i, y_i, t_i, p_i)$, where $x_i$, $y_i$ denote the pixel coordinate of event; $t_i$ denotes the time stamp; $p_i$ denotes the polarity, referring to the increase or decrease of light intensity, which is usually expressed by 1 or 0. In this paper, because the same edge will generate events with different polarities when it moves in opposite directions, the polarity should be ignored and both kinds of events should be treated equally.

In order to acquire enough information to extract and track features, a fixed number of events or events within a fixed time interval need to be aggregated along time dimension, which means squeezing time dimension, resulting in an event slice. In this paper, a fixed number is more suitable, because the speed of the camera varies widely, causing a huge difference in the number of events generated in a fixed time interval, which makes the aggregation result unstable. The aggregation process in this paper can be expressed in Equation 1.

$$es = sign\left(\sum e_i\right), \quad i_0 \leq i < i_0 + N \quad (1)$$

where $es$ denotes the event slice; $sign(\cdot)$ is defined as: $sign(x) = 1, if\ x > 0; sign(x) = 0, if\ x = 0$; $N$ is the number of events aggregated in an event slice. In summary, the pixel is set to 1 if it produces any events, otherwise set to 0. The visualization of event slices shows in Figure 1.

### B. Superimpose event slices

In order to fuse event slices with original intensity images, they need to be aligned in time in advance. The event slice aggregated from the moment an image is taken is called its corresponding event slice, which is then weighted and superimposed on the intensity image. Because an event slice contains too much noise, it is smoothed by a *Gaussian* filter before superimposition. Besides, in order to avoid features in original images are destroyed by event slices, a threshold is set to limit the superimposition. The final fusion result can be expressed in Equation 2.

$$\begin{cases} EAS(x,y) = I(x,y) + \alpha \times G(es(x,y)), & I(x,y) < \beta \\ EAS(x,y) = I(x,y), & I(x,y) \geq \beta \end{cases} \quad (2)$$

where $EAS$ is the final fusion result; $I$ is the original intensity image; $\alpha$ denotes the weight, and the larger $\alpha$, the greater the impact of event slices on fusion results; $G(\cdot)$ denotes *Gaussian* filter function; $\beta$ is the threshold limiting superposition, and only pixels with the value smaller than $\beta$ can be enhanced by the corresponding event slice.

Regarding the wight $\alpha$, an adaptive strategy is introduced: $\alpha$ is set to the maximum pixel value of the intensity image, and a lower bound is set for it as well. The whole data fusion algorithm presents in Algorithm 1.

---
**Algorithm 1** Events Aggregation and Superimposition
---
**Input:** I(intensity image), e(event stream)
**Output:** EAS(fusion result)
1: **for** $i$ in $[i_0, i_0 + N)$ **do**
2:     $es(e_i.x, e_i.y) \leftarrow 1$
3: **end for**
4: $es \leftarrow Gaussian(es)$
5: $\alpha \leftarrow max(I, \gamma)$
6: **for** $p(x,y)$ in $I$ **do**
7:     **if** $p(x,y) < \beta$ **then**
8:       $EAS(x,y) \leftarrow p(x,y) + \alpha * es(x,y)$
9:     **else**
10:      $EAS(x,y) \leftarrow p(x,y)$
11:    **end if**
12: **end for**
13: **return** EAS

---

## 4. Experiments

### 4.1. Visualization of Sources

In the experiment, the fusion results (i.e., EAS) are input as a data source into two ego-motion estimation frameworks respectively, and the performance of estimation is evaluated on a public dataset. The dataset was built through a DAVIS that produces intensity images and event stream simultaneously, hence they were already temporally aligned. Two sequences acquired in difficult illumination conditions were involved in the experiment: *hdr_boxes* and *hdr_poster*.

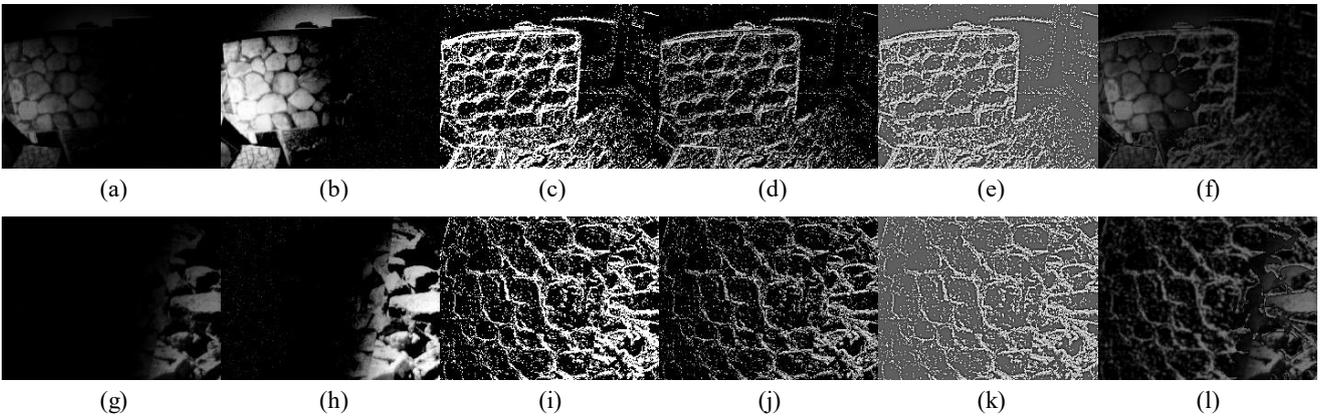

Figure 1: Original images, enhanced images, some representations of events and EAS in the dataset. (a)~(f) are from *hdr_boxes*; (g)~(l) are from *hdr_poster*. (a) and (g) are original images; (b) and (h) are enhanced images; (c) and (i) are event slices; (d) and (j) are TS; (e) and (k) are SITS; (f) and (l) are EAS. Intuitively, EAS has the best image quality.

Then the estimation accuracy of EAS was compared with that using other sources, including original images, enhanced images, event slices, TS and SITS. All these sources are visualized in Figure 1.

Because original images are underexposed and their contrast is excessively low, histogram equalization is imposed to obtain enhanced images. However, although intensity of whole images increases as well as the contrast, a large region in the images is still black, where the light is too dim and can't be recorded in traditional intensity images. For event slices, TS and SITS, they are able to acquire the information of darker environment due to the higher dynamic range of event cameras, but they all contain too much noise, causing the instability in feature extraction, tracking and matching. For EAS, it combines intensity images with processed event slices to acquire more information while reducing noise. Compared with original images and the representations of events, EAS has much higher image quality.

### 4.2. 3-DoF Rotation Estimation

In this section, we build a 3-DoF rotation estimation framework based on optical flow tracking, and apply different sources as input image sequences to estimate the 3-DoF rotation of the camera respectively. The framework is demonstrated in Figure 2.

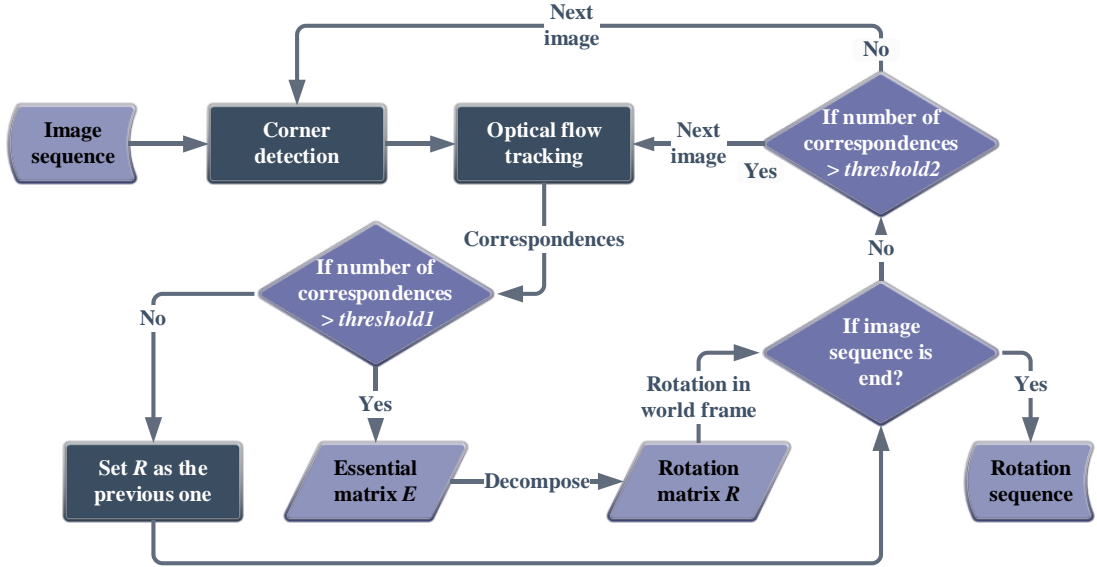

Figure 2: Diagram of the 3-DoF rotation estimation framework. An image sequence is input into the framework, and the corresponding rotation matrix sequence is output.

In the framework, *Shi-Tomasi* corners are detected and tracked by calculating LK optical flow. In this way, corresponding points in adjacent images can be found. Then the essential matrix $E$ is computed based on five-point algorithm and decomposed using SVD to get the rotation matrix $R$ relative to the previous pose. The current rotation in world frame can be obtained by multiplying $R$ by the previous rotation matrix. Because the accuracy of $R$ will decrease when there are too few correspondences, *threshold1* and *threshold2* are introduced into the process. If the number of correspondences is less than *threshold1*, current rotation matrix will be set the same as the previous one. In addition, if it is less than *threshold2*, corners will be detected again in the next image. The estimation framework doesn't contain any optimization part, and the estimation accuracy mainly depends on the number of correct correspondences.

In the experiment, about the first 25 seconds of both sequences are input into the framework to compute the corresponding rotation matrix sequences. The estimation results and ground truth show in form of *Euler* angles in Figure 2. We apply average APE to evaluate the estimation accuracy, which is calculated by Equation 3.

$$APE_{average} = \frac{1}{N} \sum_{i=1}^{N} \| log(T_{gt,i}^{-1} T_{esti,i})^{\vee} \|_2 \qquad (3)$$

In this case, transform matrix $T$ is the rotation matrix without translation component. The average APE of estimated rotations using different sources is listed in Table 1, as well as the average number of correspondences (NC).

| Sequence | *hdr_boxes* | | *hdr_poster* | |
|---|---|---|---|---|
| Source Type | average NC | average APE | average NC | average APE |
| Original | 22 | 0.69 | 33 | 0.66 |
| Enhanced | 24 | 0.39 | 58 | 0.58 |
| Event slice | 344 | 0.39 | 293 | 0.32 |
| TS | 267 | 0.41 | 238 | 0.30 |
| SITS | 320 | 0.35 | 272 | 0.32 |
| EAS | 62 | 0.17 | 93 | 0.25 |

Table 1: The average number of correspondences and APE in 3-DoF rotation estimation through different sources.

According to Table 1, *Shi-Tomasi* corners are detected and tracked worst in original images sequence, thus its average APE is the highest. The number of correspondences increases when images are enhanced by histogram equalization, and the average APE decreases accordingly. For all representations of events including event slice, TS and SITS, correspondences are much more than those in intensity images, while their average APEs don't greatly reduce accordingly. The reason is that the quality of optical flow tracking is not good and there are many wrong correspondences, caused by too much noise in them. For EAS, it achieves the best performance in both sequences, where the correspondences tracked are more than those in original images reasonably and the average APE decrease significantly. Compared with original images, its average number of correspondences has roughly tripled, and its average APE decreases by 69%. Therefore, EAS is able to improve the quality of optical flow tracking of *Shi-Tomasi* corners in nonideal lighting conditions effectively.

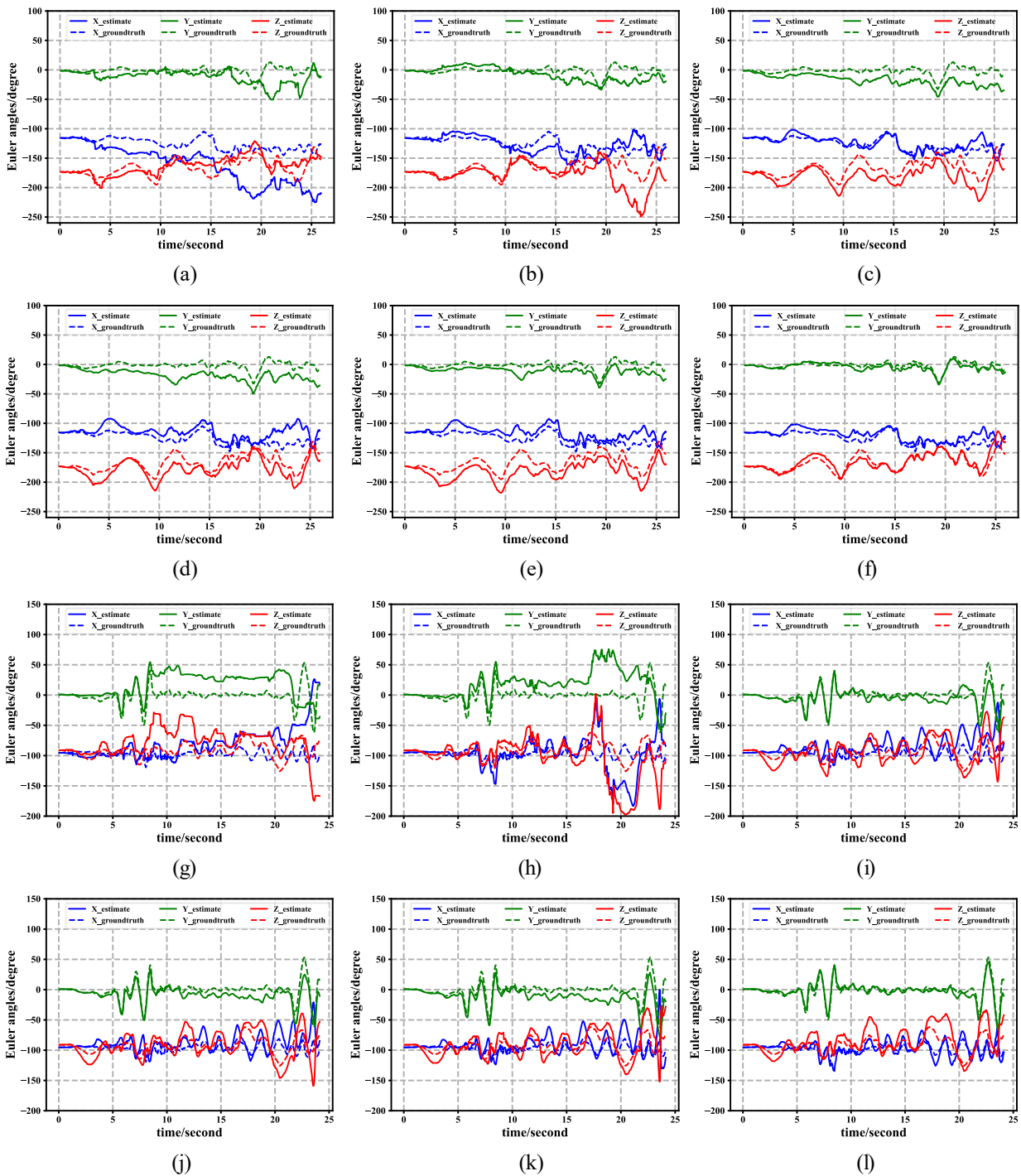

Figure 3: *Euler* angles of estimated rotations and ground truth in the dataset. (a)~(f) are from *hdr_boxes*; (g)~(l) are from *hdr_poster*. (a) and (g) are estimated through original images; (b) and (h) are estimated through enhanced images; (c) and (i) are estimated through event slices; (d) and (j) are estimated through TS; (e) and (k) are estimated through SITS; (f) and (l) are estimated through EAS. The estimation result of EAS fits the ground truth best.

### 4.3. 6-DoF Pose Estimation

In this section, original images and EAS are carried as input to the monocular ORB-SLAM2 to estimate the 6-DoF pose of the camera respectively. ORB-SLAM2 detects ORB features in each image and the corresponding points are found by comparing their descriptors. If too few key-points are correctly matched, the system will lose the track of the camera. The poses and APEs of keyframes recorded by ORB-SLAM2 using different sources show in Figure 4, and the average APE is listed in Table 2.

| Sequence | *hdr_boxes* | | *hdr_poster* | |
|---|---|---|---|---|
| Source Type | Original | EAS | Original | EAS |
| average APE | 0.021 | 0.029 | 0.049 | 0.058 |
| ORB features | 199 | 344 | 212 | 392 |
| matches | 94 | 57 | 114 | 53 |

Table 2: The average APE of recorded keyframes through different sources, and the average number of ORB features and matches in different sources.

According to Table 2, although EAS contains more information than original images in dim environment, ORB-SLAM2 doesn't works better. For estimation accuracy, the average APE of using EAS is a little higher than using original images. We analyze the reasons from the perspective of feature extraction and matching, and the average number of extracted ORB features in each image and the average number of matched key-points in adjacent images are both listed in Table 2. In fact, although much more ORB features can be detected in EAS, the matched key-points reduce instead, which means ORB features in EAS are unstable and not very suitable for matching. Therefore, EAS cannot improve the performance of ORB-SALM2 in difficult lighting conditions, but has a negative impact.

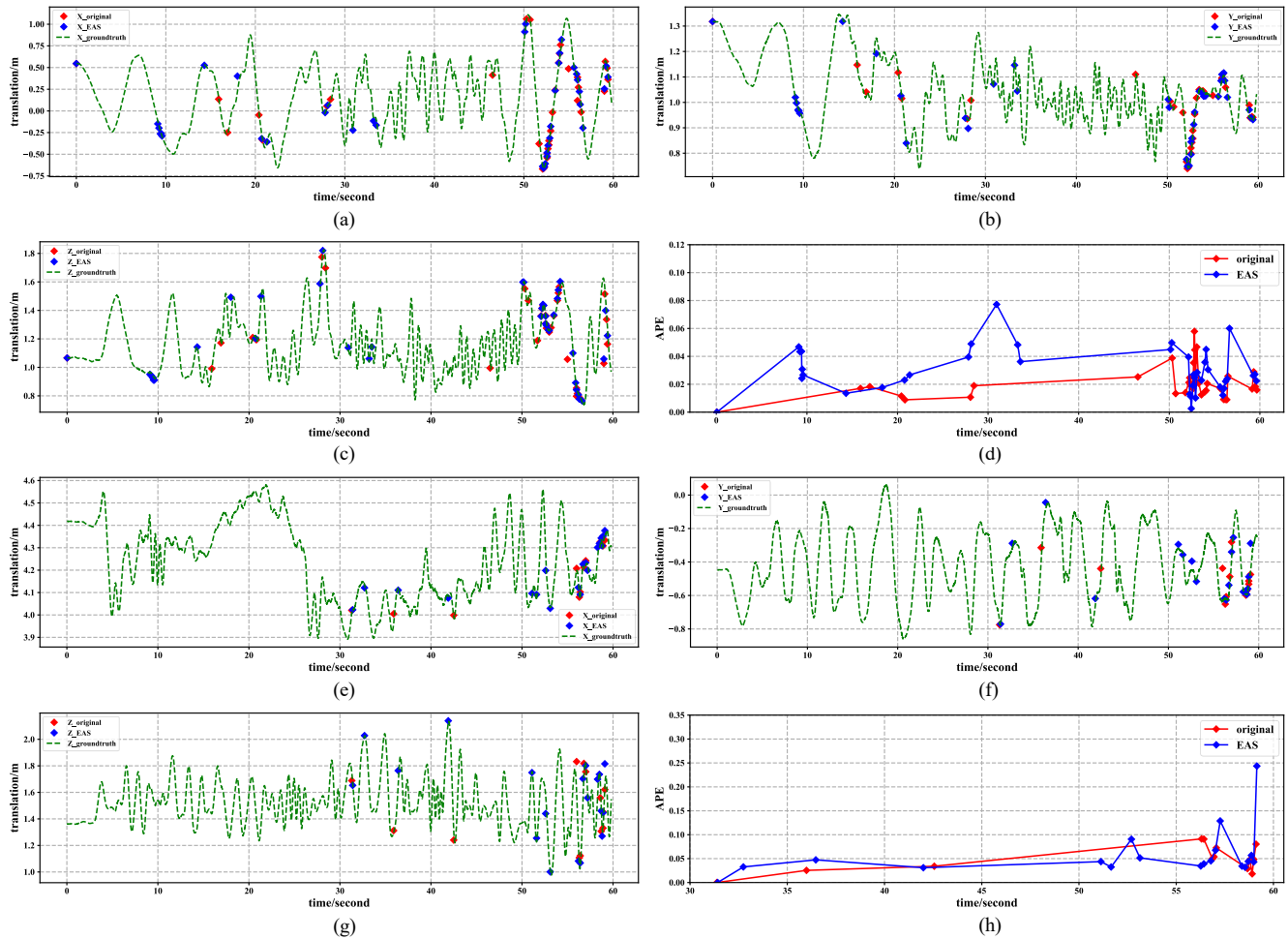

Figure 4: Poses and APEs of the keyframes recorded through original images and EAS. (a)~(d) are from *hdr_boxes*; (e)~(h) are from *hdr_poster*. (a)~(c) and (e)~(g) show the translations of keyframes in the whole trajectories of the camera; (d) and (h) are APEs of the keyframes. In this case, EAS doesn't improve the estimation accuracy.

## 5. Conclusions

In this paper, we propose a novel data fusion algorithm called EAS to fuse conventional intensity images with the event stream output by event cameras. The fusion result is applied to two ego-motion estimation tasks: 3-DoF rotation estimation based on optical flow tracking and 6-DoF pose estimation using ORB-SLAM2, and it is evaluated based on a public dataset acquired in dim scenes. In our rotation estimation framework, EAS reduces the average APE by 69% compared to original images, while it does not work well in ORB-SLAM2. The result shows that our algorithm is able to effectively leverages the high dynamic range of event cameras to improve the performance of the ego-motion estimation framework based on optical flow tracking in difficult illumination conditions. In future work, we will try more SLAM frameworks and further improve our algorithm to fit them better.

**Acknowledgments**

Thanks to Nanjing Institute of Intelligent Technology for providing relevant experimental platforms and technical support.